\pdfoutput=1

\documentclass[11pt]{article}

\usepackage[final]{acl}

\usepackage{times}
\usepackage{latexsym}

\usepackage[T1]{fontenc}

\usepackage[utf8]{inputenc}

\usepackage{microtype}

\usepackage{inconsolata}

\usepackage{graphicx}
\usepackage{makecell}
\usepackage{multirow}
\usepackage{xcolor}         
\usepackage{tcolorbox}
\usepackage{hhline}

\definecolor{c1}{HTML}{801dae}
\definecolor{c2}{HTML}{008000}
\definecolor{c3}{HTML}{00BFFF}
\definecolor{c4}{HTML}{0c8918}

\newcommand{\figref}[1]{Figure \ref{#1}}

\title{A Cognitive Evaluation Benchmark of Image Reasoning and Description for Large Vision-Language Models}

\author{Xiujie Song\textsuperscript{1}, Mengyue Wu\textsuperscript{1}\thanks{Corresponding authors.}, Kenny Q. Zhu\textsuperscript{2}\footnotemark[1], Chunhao Zhang\textsuperscript{1}, Yanyi Chen\textsuperscript{3}\\
\textsuperscript{1} 
X-LANCE Lab, Department of Computer Science and Engineering \\
 MoE Key Lab of Artificial Intelligence, AI Institute \\
  Shanghai Jiao Tong University, Shanghai, China \\
  \textsuperscript{2}University of Texas at Arlington, Arlington, Texas, USA \\
  \textsuperscript{3}University of Chicago, Chicago, Illinois, USA \\
  \texttt{\textsuperscript{1}\{xiujiesong, mengyuewu\}@sjtu.edu.cn}, 
   \texttt{\textsuperscript{2}kenny.zhu@uta.edu}
}

\begin{document}
\maketitle

\begin{abstract}
    Large Vision-Language Models (LVLMs), despite their recent success, are hardly comprehensively tested for their cognitive abilities.
    Inspired by the prevalent use of the Cookie Theft task in human cognitive tests, 
    we propose a novel evaluation benchmark to evaluate high-level cognitive abilities of LVLMs using images with rich semantics.
    The benchmark consists of 251 images along with comprehensive annotations. 
    It defines eight reasoning capabilities and comprises an image description task and a visual question answering task. 
    Our evaluation of well-known LVLMs shows that there is still a significant gap in cognitive abilities between LVLMs and humans\footnote{
    Our code and data are available at:
    \url{https://github.com/xiujiesong/CogBench}.
    }. 
\end{abstract}

\section{Introduction}
\label{sec:intro}

Recently, with the emergence of Large Language Models (LLMs) such as GPT-4 \cite{openai2023gpt4}, the cognitive abilities of language models have reached a new level \cite{zhuang2023efficiently}.
They demonstrate remarkable performance in many tasks \cite{bubeck2023sparks}.
In Vision-Language (VL), several researchers \cite{zhu2023minigpt, liu2023llava, ye2023mplug} endeavor to boost Vision-Language Pre-trained Models (VLPMs) by integrating powerful LLMs \cite{touvron2023llama, chiang2023vicuna}, referred to as Large Vision-Language Models (LVLMs)~\cite{li-etal-2023-evaluating}.
With LLM serving as the ``brain'', the cognitive abilities of LVLMs are enhanced, enabling them to tackle more challenging tasks~\cite{yang2023dawn}.
Some state-of-the-art LVLMs, such as GPT-4o~\cite{openai2023gpt4}, are progressing toward human-level cognitive abilities. 
Thus, there is growing interest in evaluating cognitive abilities of LVLMs.
Though some LVLM evaluation benchmarks, such as MME \cite{fu2023mme}, MMBench \cite{liu2023mmbench}, and SEED Bench \cite{li2023seed}, 
evaluate cognitive reasoning ability as one aspect of their evaluation, they do not provide a comprehensive evaluation of higher-level reasoning abilities.
Most of the images they use contain less semantics and thus require relatively little reasoning to understand.

\begin{figure}[ht]
\centering
\includegraphics[width=0.47\textwidth]{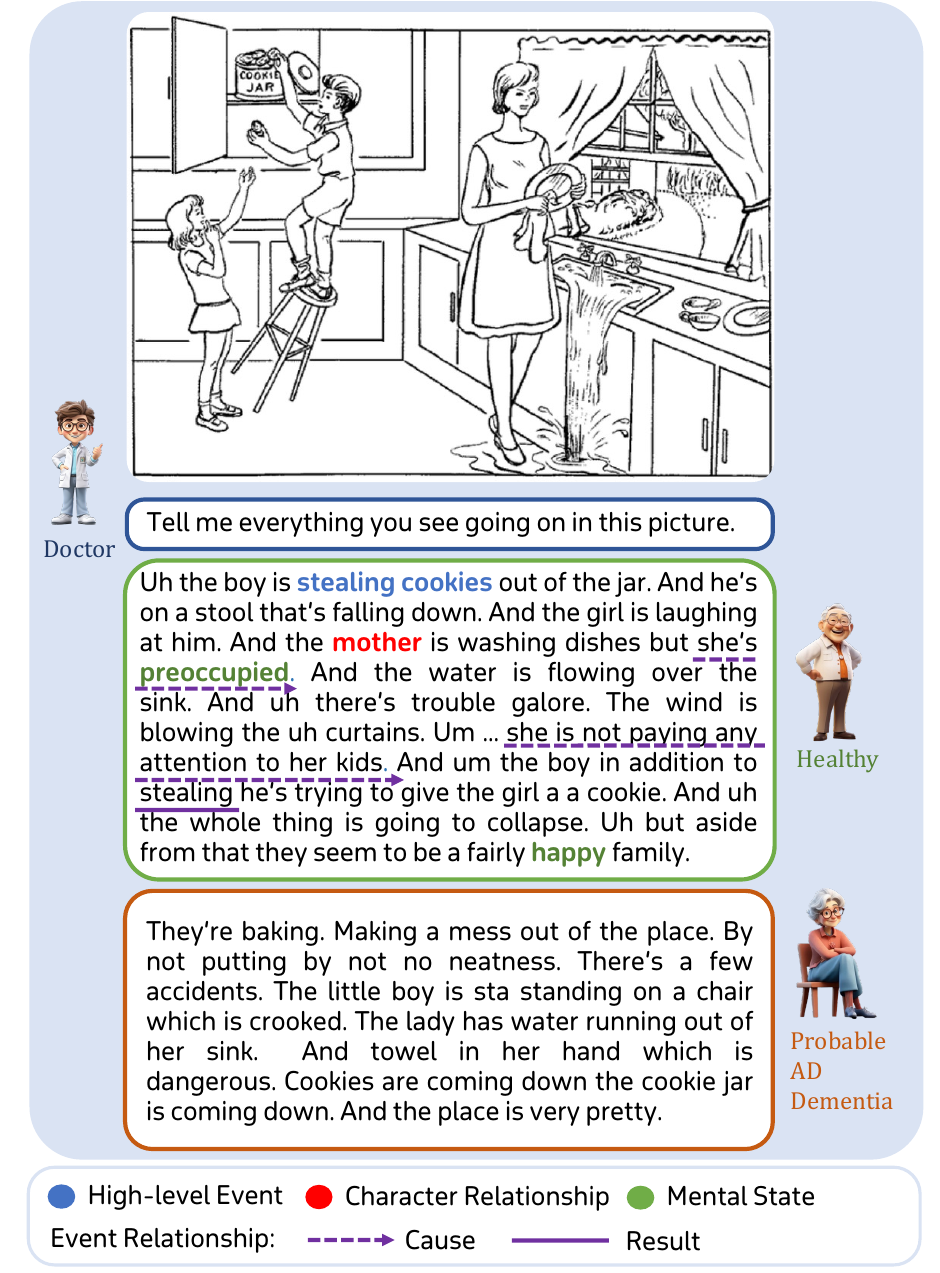}
\caption{Cookie Theft picture description task. 
    The descriptions in the green frame and the orange frame were respectively produced by a healthy 75-year-old man and a 66-year-old woman with probable AD dementia\protect\footnotemark. 
    }
\label{fig:cookie_theft}
\end{figure}

\footnotetext{The two samples are extracted from DementiaBank (\url{https://dementia.talkbank.org/}), which records description transcripts from both healthy subjects and dementia patients.}

In our study, we draw inspiration from the Cookie Theft picture description task (Figure \ref{fig:cookie_theft}), a key component of the Boston Diagnostic Aphasia Examination \cite{goodglass2001-ej}, which is widely used in clinical practice within speech-language pathology for \textbf{language and cognitive function screening}~\cite{describe-ctp, mueller2018connected}. 
Notably, despite being designed more than half a century ago, this picture remains prevalent in contemporary psychological discussions.

Is it possible to transfer the success of the Cookie Theft picture description in human cognitive tests to evaluating the cognitive abilities of LVLMs? 
Linguists and psychologists~\citep{describe-ctp} conducted an analysis to determine the factors contributing to the success of the Cookie Theft picture. 
The study reveals that the narrative includes information with varying levels of importance and encompasses a broad range of semantic categories. 
It is observed that during the description of the Cookie Theft picture, individuals with intact cognitive functions exhibit their cognitive prowess by logically deducing the events and their interconnections, the relationships between characters and their mental states, etc. 
In contrast, those with cognitive impairments tend to merely list the superficial aspects of the situation without deeper reasoning.
For instance, in Figure~\ref{fig:cookie_theft}, by comparing the descriptions produced by the healthy man and the woman with probable Alzheimer's disease (AD) dementia, we can identify the following differences:

\begin{itemize}
    \item The description produced by the healthy man used ``\textbf{mother}'' instead of ``lady'', indicating reasoning about \textbf{character relationship}. 
    \item The healthy man used ``\textbf{stealing cookies}'' instead of ``taking cookies'', indicating his reasoning about this \textbf{high-level event}. 
    The description produced by the patient even did not mention this event at all. 
    \item The healthy man used ``the mother is \textbf{preoccupied}'' and ``\textbf{happy}'' to describe people's \textbf{mental state}. 
    \item The description reflected the  \textbf{causal relationships between events}. ``\textbf{The kids are stealing cookies}'', because ``\textbf{the mother is preoccupied}'' and ``\textbf{not paying attention to her kids}''. 
\end{itemize}

Through these reasoning processes, the difference in cognitive abilities between the two individuals is reflected in their descriptions. 
A picture that can evaluate cognitive functions needs to be carefully designed and crafted.
\citet{tasnim-etal-2022-depac} introduced guidelines for drawing pictures similar to Cookie Theft, which is consistent with the findings mentioned above.
Generally speaking, compared to ordinary images, Cookie Theft-like images feature i) \textit{a prominent story theme}, ii) \textit{richer content}, iii) \textit{display complex relationships among entities}, and thus require stronger cognitive abilities to understand and describe.

With the above design principles, we propose to construct a \textbf{Cog}nitive Evaluation \textbf{Bench}mark, referred to as \textbf{CogBench}, to evaluate cognitive abilities of LVLMs mainly from the reasoning perspective using high-quality Cookie Theft-like images.
CogBench defines eight core cognitive reasoning capabilities, including reasoning about special time, location, character, character relationship, event, event relationship, next moment event and mental state. 
Both a generative Image Description task and a discriminative Visual Question Answering (VQA) task are designed.
Our main contributions are as follows:

\begin{itemize}
	\item To the best of our knowledge, this is the first-of-its-kind attempt to incorporate the concept of the well-known Cookie Theft picture description task, originally designed for human cognitive testing, into the cognitive evaluation of LVLMs.
	\item Taking insights from human cognition research, we are the first to define Cookie Theft-like images with eight reasoning dimensions and to create a dataset with semantically complex images aligned with the Cookie Theft. This approach allows for a comprehensive evaluation of the visual reasoning capabilities of LVLMs across these dimensions.

	\item Our evaluation on existing LVLMs shows that a significant gap exists between the cognitive abilities of LVLMs and human beings, indicating CogBench will be a valuable evaluation benchmark in the near future.

\end{itemize}

\section{Dataset Construction}

In this section, we will introduce the construction of CogBench, detailing its image collection, annotation process, tasks, and data statistics.

\subsection{Image Collection}

\begin{figure}[htbp]

\centering
    \centering
    \includegraphics[width=0.45\textwidth]{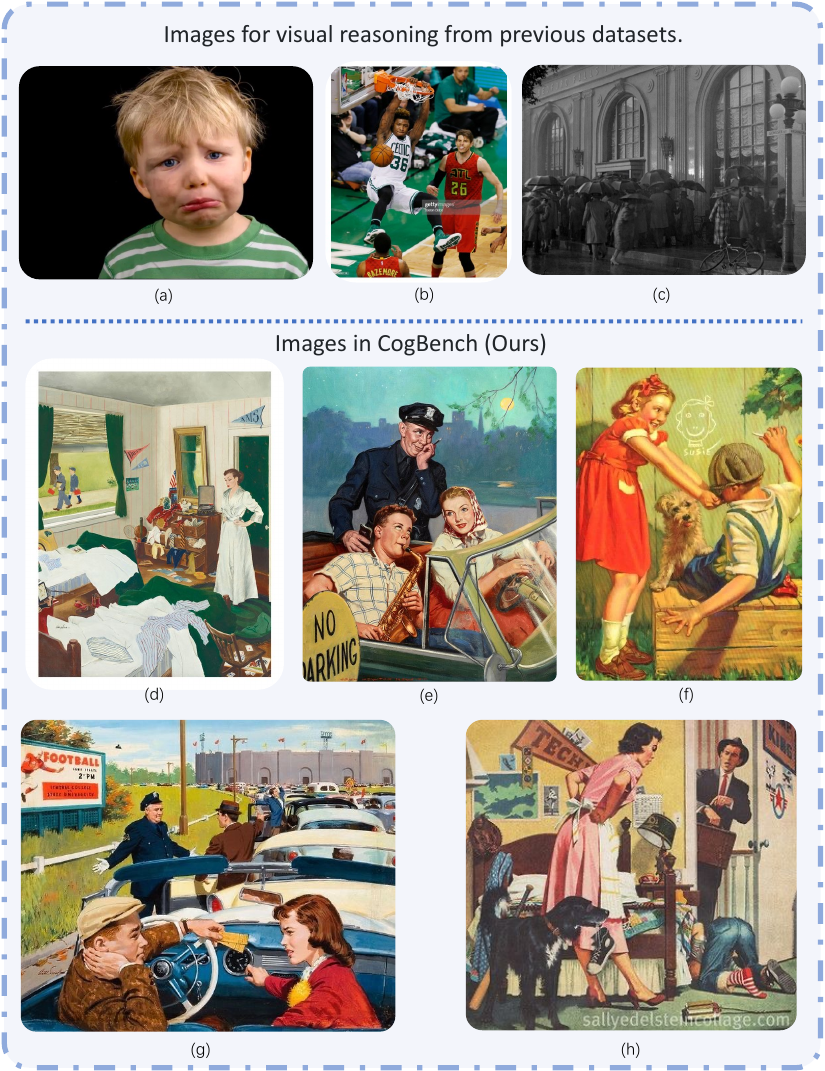}
    \caption{Comparison between our images and those from previous visual reasoning tasks.
    Our images contain rich entities and CoRs. 
    Compared to our images, image (a) has fewer entities and CoRs, while image (b) and (c) have some entities but fewer CoRs.
    }
    \label{fig:cogbench_example}
\end{figure}

Building on previous studies \cite{describe-ctp, tasnim-etal-2022-depac}, we establish the following criteria for collecting Cookie Theft-like images that we propose: 

\paragraph{Rule 1: Storytelling} The image depicts an interesting story. For instance, the Cookie Theft picture tells the story of a mother busy washing dishes while two kids take the opportunity to stand on a stool and sneakily steal cookies.  
\paragraph{Rule 2: Rich Chain-of-Reasonings} Images should display rich Chain-of-Reasonings (\underline{\textbf{CoR}s}) in a scene. 
 A CoR connects low-level observations in an image to produce a high-level reasoning conclusion or connects the cause and effect of events.
For example, ``The mother is busy washing dishes. + The boy is standing on the stool behind the mother. + The girl standing by the boy is shushing him. + The boy is fetching cookies from the jar in the cabinet. $\rightarrow$ The boy and girl are stealing cookies.'' is a CoR about the high-level event ``stealing cookies''. 
Note that a story is usually constructed through various CoRs.
\paragraph{Rule 3: Restricted Content Complexity} Images should contain rich content but not be overly complex. The number of entities should be sufficient to support a good story while being restricted to emphasize the key points effectively.

With the above criteria, 
we manually collect pictures from Pinterest\footnote{\url{https://www.pinterest.com/}}, 
and the Cookie Theft picture is also included. 
\figref{fig:cogbench_example} shows the differences between our images and those from other datasets by examples.
Note that most of the images in CogBench are in a painting style because they are abstracted from reality and tend to contain richer CoRs than real-world images, making them more effective for picture description tasks. 
Even so, painting-style images that meet our criteria are still rare, highlighting the value of our dataset.

\subsection{Image Annotation}
\label{sec:annotation}

\begin{figure*}[h]
  \centering
  \includegraphics[width=0.939\textwidth]{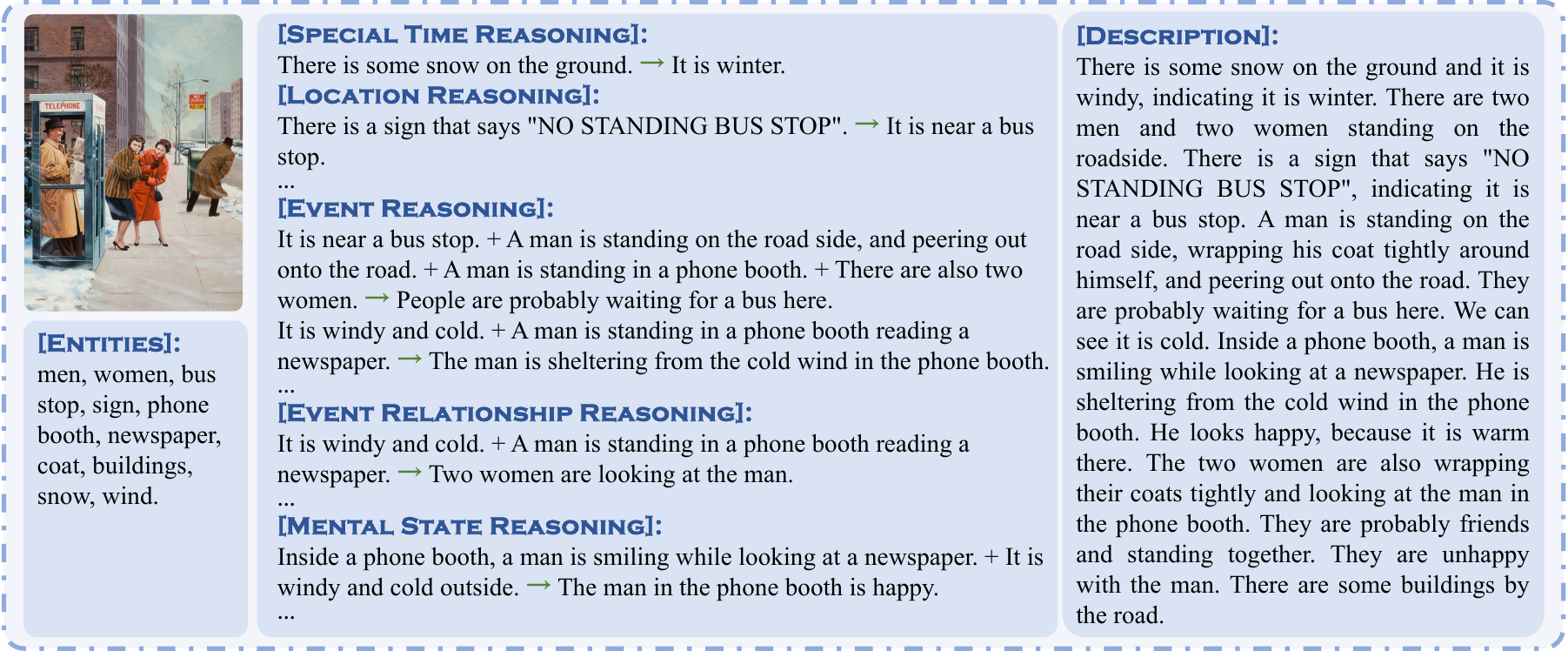}
  \caption{An example of the Description task from CogBench. 
}
  \label{fig:cogbench}
\end{figure*}

Human annotators, mostly undergraduate or graduate students aged 18-28, are hired to annotate the collected images.
As shown in Figure \ref{fig:cogbench}, the annotation includes three parts: [Entities], [CoRs] and [Description].
By annotating [Entities] and [CoRs], we aim to evaluate the low-level recognition ability and high-level cognitive reasoning ability of models respectively based on their descriptions. 
[Description] is annotated as the reference description for the image.
The three parts are annotated in that order.

\noindent$[$\textbf{Entity Annotation}$]$ We ask annotators to list as many entities in the image as possible and entities
that are difficult to recognize should be omitted. 

\noindent$[$\textbf{CoR Annotation}$]$ 
To evaluate model cognition in a fine-grained manner, we design eight reasoning dimensions based on studies in human cognition~\cite{describe-ctp, Byom2013-BYOTOM, ADDIS20071363} and Computer Vision (CV)~\cite{FZCVR22, park2020visualcomet, zellers2019vcr, MEmoR2020}.
CoRs for these dimensions are annotated:
\begin{itemize}
    \item \textbf{Special Time Reasoning}: reasoning about the special time of the story in the image, e.g., festivals.
  \item \textbf{Location Reasoning}: reasoning about the location of the story in the image, e.g., near a school.
  \item \textbf{Character Reasoning}: reasoning about the characters of subjects in the image, e.g., a doctor.
  \item \textbf{Character Relationship Reasoning}: reasoning about the relationships between characters in the image, e.g., ``the woman is the mother of the kids.''
  \item \textbf{Event Reasoning}: reasoning about the high-level events in the current and previous moments in the image. 
  The difference between high-level and low-level events lies in the amount of semantic information they contain. 
  For example, ``stealing cookies'' is a higher-level event compared to ``taking cookies'' as it additionally conveys the semantic of ``taking advantage without permission or knowledge.''
  \item \textbf{Event Relationship Reasoning}: reasoning about the causal and temporal relationships between different events in the image. For instance, ``the sink is overflowing \textbf{because} the mother left the tap on.''  
  \item \textbf{Next Moment Event Reasoning}: reasoning about the events that will happen in the next moment in the image. 
  For example, ``the police officer will reprimand the boy who violates the rules.''
  \item \textbf{Mental State Reasoning}: reasoning about the mental states of subjects in the image, including their emotions, thoughts, and other psychological states. For example, ``the girl appears to be happy.''

\end{itemize}

\noindent$[$\textbf{Description Summary}$]$ Annotators are finally asked to write a description that conveys the entire story in the image based on the annotated [Entities] and [CoRs].

The complete annotation instruction can be found in Appendix \ref{sec:annotation_instruction}. 
Considering different people may have different understanding about some images, we ask three annotators to annotate each image. 
Then, we draw on the idea of majority voting to merge the three annotations into one.
For [Entities] and [CoRs], we first accept most of the entities and CoRs that are annotated by at least two annotators. 
Other entities and CoRs are also included if reasonable.
The final [Description] is obtained by modifying the best annotated [Description] with the merged [Entities] and [CoRs]. 
We discard images where there is a significant difference in understanding among the three annotators.

\subsection{Tasks in CogBench}

We design a generative Image Description task and a discriminative Multiple-Choice Question Answering task in CogBench.

\subsubsection{Image Description Task}
This is the primary task of the benchmark.
The difference between our description task and existing image description tasks~\cite{xie2022visual, zhu2023chatgpt, zhuge2023mindstorms} is that we expect LVLMs to understand and describe the story in the image through high-level cognitive reasoning. 
For instance, in Figure \ref{fig:cogbench}, the description of the image should not only include what is in the picture but also focus on elucidating the story of ``on a cold winter day, a man is reading a newspaper in a phone booth near the bus stop to escape the cold, while two passing women express their displeasure upon seeing that'' through a series of reasoning processes.

\subsubsection{Visual Question Answering Task}

\begin{figure}
  \centering
  \includegraphics[width=0.465\textwidth]{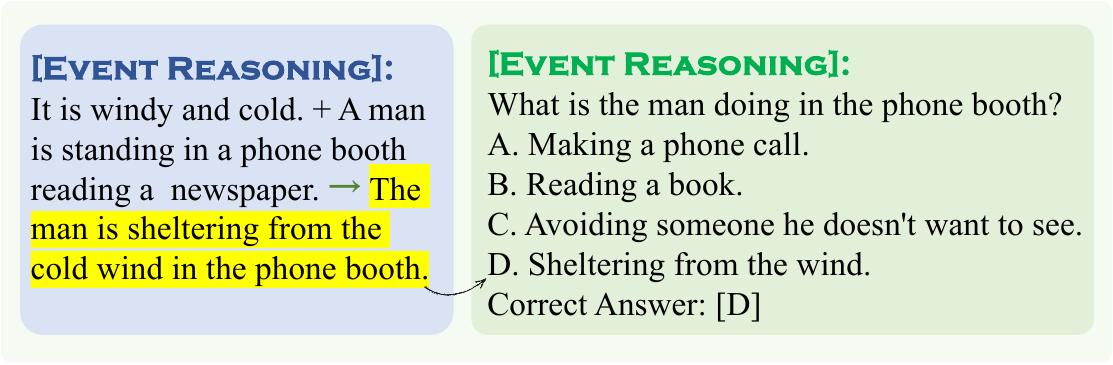}
  \caption{Generating a multiple-choice question based on an [Event Reasoning] CoR annotation.}
  \label{fig:qa_example}
\end{figure}

The VQA task features standard four-option multiple-choice questions, easing the evaluation process. Like the Description task, VQA questions involve different types of high-level cognitive reasoning, as illustrated by the question about event in Figure \ref{fig:qa_example}. 
We use GPT-4 to assist in generating questions based on the annotations from Section \ref{sec:annotation}. 
With the annotated CoRs, both the conclusion (right side of$\rightarrow$) and the reasoning behind it (left side of $\rightarrow$) in each CoR can be used to generate questions and corresponding options, as depicted in Figure \ref{fig:qa_example}.
These components in each CoR provide the correct options directly to the questions generated based on them. 
Specifically, this process is two-fold. 
1) \textbf{Automated Question Generation}: We use GPT-4 to generate questions for CogBench images, tailoring prompts for each reasoning category to produce CoR-related questions. The key point is to prompt GPT-4 to generate higher-quality distractors. An example prompt for this CoR-based GPT-assisted question generation approach is provided in Appendix \ref{sec:qa_prompt}. 
2) \textbf{Manual Refinement}: 
Despite GPT-4's capabilities, some generated questions are not challenging enough. 
In this stage, we manually refine the questions, ensuring they do not overtly favor the correct answer and that distractors are closely related to the question and misleading. 
Additionally, ChatGPT aids in identifying and filtering out simple questions that can be answered without image input.

\subsection{CogBench Statistics}

\begin{table*}[th]
  \centering
  \small
  \setlength{\tabcolsep}{2.5pt} 

  \begin{tabular}{lcccccccc}
  \hline
  \textbf{} & \thead{\textbf{Time} } & \thead{\textbf{Location} } & \thead{\textbf{Character} } &  \thead{\textbf{Character}\\ \textbf{Relationship} } & \thead{\textbf{Event} } & \thead{\textbf{Event} \\ \textbf{Relationship} } & \thead{\textbf{Next Moment} \\ \textbf{Event} } & \thead{\textbf{Mental State} } \\ 
  \hline
  CoR & 47 & 177 & 106 & 263 &  701 & 425  & 107 & 417 \\
  QA & 86  & 220 & 162 & 317 & 658   &  402  & 135 & 597 \\  
  \hline
  \end{tabular}
  \caption{\label{tab:stat}
  Distribution of CoRs and questions in CogBench.
  }
\end{table*}

CogBench consists of 251 semantically rich images with a total of 2670 entities, 2243 CoRs, 251 descriptions and 2577 questions, indicating the content contained in each image is complex, showcased in Table \ref{tab:stat}. 
The number of CoRs of event-related reasoning and [Mental State Reasoning] is large, which is a manifestation of the rich interesting stories in the images.

\section{Experiments}
\label{sec:experiments}

We will evaluate the selected LVLMs, detail evaluation strategies and discuss the results in this section. 

\subsection{Large Vision-Language Models}
\label{sec:lvlm}
We evaluate a selection of recent representative open-source and closed-source LVLMs, including InstructBLIP-7B \cite{dai2023instructblip}, Qwen-VL series models~\cite{bai2023qwenvl, Qwen2VL}, mPLUG-Owl-2 \cite{ye2023mplug}, LLaVA series models~\cite{liu2023improved, liu2024llavanext, li2024llava}, ShareGPT4V \cite{chen2023sharegpt4v}, CogVLM series models~\cite{wang2023cogvlm, hong2024cogvlm2}, InternVL2-26B~\cite{chen2023internvl}, GPT-4V and GPT-4o \cite{openai2023gpt4}. 
A brief introduction to these models is provided in Appendix \ref{sec:lvlms}.

\subsection{CogBench Evaluation Strategy}

\subsubsection{Evaluation of the Description Task}

\paragraph{Evaluation Modes}
For the Description task, we set up two evaluation modes: Spontaneous Description and Directed Reasoning. 
In the Spontaneous Description mode, we prompt the LVLMs with the following instruction to obtain detailed image descriptions: 
``\texttt{Describe this image in detail.}''
This mode is more similar to the Cookie Theft picture description task, which aims to stimulate spontaneous descriptions~\cite{MATIASGUIU2022141}. 
It can help analyze the behavior of LVLMs when they describe images.
For the Directed Reasoning mode, the corresponding prompt is:
``\texttt{Please provide a detailed description of the story depicted in the image, including high-level reasoning about the time and location, the roles and relationships of the characters, the events and their causal relationships, what might happen next, and the mental states of the characters.}''
This mode simplifies the Description task compared to the Spontaneous Description mode and aims to evaluate whether models can reason correctly when they know the directions we expect.

\paragraph{Evaluation Metrics}
We consider model performance at two levels: low-level Recognition ability and high-level Cognition ability. 
Evaluation metrics for both levels are calculated based on recall scores, referred to as Recognition Score and Cognition Score, respectively.

The \textbf{Recognition Score} is calculated as the ratio of recognized [Entities] to annotated [Entities] across all images.
First, we use spaCy\footnote{\url{https://spacy.io/}} to extract nouns from the model-generated description, and use sentence-transformers\footnote{\url{https://www.sbert.net}. The model \textit{all-mpnet-base-v2} is adopted.} to encode the annotated [Entities] and extracted nouns into embeddings. 
Then, we calculate the cosine similarity between the embeddings of the [Entities] and the nouns. 
For each entity, if the cosine similarity score between the entity and any noun is greater than a threshold (0.6 in this paper), we consider the entity to be recognized by the model.

For the \textbf{Cognition Score}, we calculate the scores for each of the eight cognitive reasoning dimensions, as well as an overall score using GPT-4. 
To enhance objectivity and granularity, GPT-4 is utilized for a binary classification task to assess if a generated description includes the semantics of each annotated CoR. 
For reasoning types other than [Event Relationship Reasoning], we task GPT-4 with determining whether the conclusion in each CoR is mentioned in the description. 
For [Event Relationship Reasoning], we task GPT-4 with determining whether each causal relationship between events (i.e., the entire CoR), as annotated, is present in the description. 
The CoR scores for each dimension are then used to compute a recall score for each respective type. 
The overall Cognition Score is the sum of all CoR scores divided by the total number of CoRs.
The corresponding prompts are shown in Appendix \ref{sec:cogid_eval_prompt}.

\subsubsection{Evaluation of the VQA Task}

For multiple-choice questions in the VQA task, we use accuracy as the evaluation metric. 
As questions are generated based on CoRs, we can also calculate the accuracy for each reasoning capability as well as the overall cognitive capability.

\subsection{Results of the Description Task}

We evaluate the LVLMs on the Description task in terms of both recognition and cognition abilities. 
As a reference, we also calculate traditional image captioning evaluation metrics by comparing the model-generated description with the annotated reference [Description], and details are shown in Appendix \ref{sec:cap_eval}.

\subsubsection{Recognition}

Table \ref{tab:rec_score} shows the Recognition Scores of models on the Description task.
InternVL2-26B, LLaVA-OV-7B-Chat, GPT-4o and Qwen2-VL achieve relatively better performance than other models, which means they can recognize and describe more entities.
GPT-4V, CogVLM and CogVLM2 also demonstrate competitive performance.
It can be concluded that the recognition ability of open-source LVLMs is approaching that of GPT-4o, with some models even surpassing it.
Nevertheless, other open-source LVLMs have a significant gap in recognition capability compared to GPT-4,
indicating some LVLMs still have room for development before reaching the recognition capability of GPT-4.
Note that some models perform worse in mode 2 than in mode 1, which could be because these models focus more on high-level semantic reasoning and overlook the description of some low-level entities.
Although many LVLMs perform well, such as InternVL2-26B and GPT-4o, they still miss many entities, suggesting room for improvement in recognition capability.

\begin{table}[h!]
    \centering
    \small 
    \setlength{\tabcolsep}{2.5pt} 
    \begin{tabular}{lcc}
    \hline
    \multirow{2}*{\thead{\textbf{Model}}} & \multicolumn{2}{c}{\thead{\textbf{Recognition Score}}} \\ 
    \cline{2-3}
        & Mode 1 & Mode 2  \\
    \hline
    InstructBLIP-7B  & 40.0 & 36.4 \\  
    Qwen-VL-Chat & 43.3 & 45.8 \\
    LLaVA-v1.5-7B  & 39.8 & 41.2 \\  
    LLaVA-v1.5-13B & 41.0 & 39.3 \\ 
    mPLUG-Owl-2 &  37.4  & 37.8 \\ 
    ShareGPT4V-7B & 46.9 & 47.3 \\ 
    ShareGPT4V-13B & 48.7 & 47.4 \\ 
    LLaVA-v1.6-vicuna-7B  & 49.3 & 51.1\\
    LLaVA-v1.6-vicuna-13B  & 53.3 & 53.9 \\
    LLaVA-v1.6-34B  & 52.2 & 52.0 \\
    CogVLM-Chat & 61.6 & 56.5 \\
    CogVLM2-Llama3-Chat & 62.3 & 58.9 \\
    InternVL2-26B & \colorbox{gray!20}{\textbf{70.7}} & 65.9 \\ 
    Qwen2-VL-7B & 66.3 & 59.6 \\
    LLaVA-OV-7B-Chat & 67.4	 & \colorbox{gray!40}{\textbf{72.3}} \\ 
    GPT-4V & 62.9 & 56.5 \\ 
    GPT-4o & 66.8 &  65.1 \\ 
    \hline
    Oracle & \multicolumn{2}{c}{92.8} \\ 
    \hline
    \end{tabular}
    \caption{
        \label{tab:rec_score}
        Recognition Scores of LVLMs on the Description task. 
        Mode 1 and Mode 2 refer to the Spontaneous Description mode and the Directed Reasoning mode respectively.
        For reference, the Recognition Score of Oracle is calculated based on the annotated [Description] in CogBench dataset as an estimated upper bound. 
        Numbers are presented in \% with a full score of 100\%.
        }
\end{table}

\begin{table*}[h!]
    \centering
    \scriptsize
    \setlength{\tabcolsep}{1pt} 

    \begin{tabular}{lccccccccc}
    \hline
    \thead{\textbf{Model}} & \thead{\textbf{Time} } & \thead{\textbf{Location} } & \thead{\textbf{Character} } &  \thead{\textbf{Character}\\ \textbf{Relationship} } & \thead{\textbf{Event} } & \thead{\textbf{Event} \\ \textbf{Relationship} } & \thead{\textbf{Next}\\ \textbf{Moment} \\ \textbf{Event} } & \thead{\textbf{Mental} \\ \textbf{State} } & \thead{\textbf{Overall} } \\ 
    \hline
    InstructBLIP-7B  & 14.9 / 17.0 & 54.2 / 59.9 & 24.5 / 29.2 & 30.0 / 41.1 & 10.3 / 9.3 & 4.9 / 5.6 & 1.9 / 6.5 & 17.3 / 22.8 & 16.7 / 19.8 \\
    Qwen-VL-Chat  & 23.4 / 25.5 & 57.1 / 58.8 & 28.3 / 31.1 & 29.3 / 50.6 & 15.1 / 19.4 & 12.7 / 13.2 & 3.7 / 9.3 & 12.0 / 25.7 & 19.3 / 26.3 \\ 
    LLaVA-v1.5-7B  & 8.5 / 19.1 & 45.2 / 59.3 & 15.1 / 18.9 & 18.3 / 40.7 & 8.8 / 8.6 & 4.9 / 5.6 & 1.9 / 1.9 & 13.2 / 21.8 & 12.8 / 18.6 \\
    LLaVA-v1.5-13B  & 12.8 / 14.9  & 48.6 / 55.9 & 17.9 / 19.8 & 24.3 / 39.5 & 9.8 / 9.3  & 4.7 / 5.9  & 3.7 / 4.7 & 16.3 / 21.3 & 15.0 / 18.5 \\
    mPLUG-Owl-2  & 6.4 / 12.8 & 48.0 / 57.1  & 23.6 / 22.6 & 21.3 / 41.8 & 8.4 / 9.4 & 4.0 / 4.5 & 1.9 / 3.7 & 13.7 / 17.3 & 13.6 / 17.9 \\
    ShareGPT4V-7B  & 19.1 / 14.9 & 60.5 / 58.8  & 20.8 / 25.5 & 22.4 / 35.0 & 10.0 / 12.6 & 4.0 / 6.4 & 2.8 / 3.7 & 15.1 / 17.5 & 15.6 / 18.8  \\
    ShareGPT4V-13B  & 23.4 / 17.0 & 57.1 / 60.5 & 23.6 / 29.2 & 26.2 / 37.6 & 12.4 / 12.4 & 6.6 / 7.3 & 2.8 / 3.7 & 14.4 / 16.1 & 17.1 / 19.3  \\
    LLaVA-v1.6-vicuna-7B  & 17.0 / 25.5 & 61.0 / 62.1 & 23.6 / 29.2 & 25.1 / 47.9 & 12.1 / 14.1 & 7.5 / 8.5 & 2.8 / 8.4 & 17.0 / 23.0 & 17.7 / 23.1 \\
    LLaVA-v1.6-vicuna-13B & 17.0 / 27.7 & 63.3 / 65.5 & 27.4 / 33.0 & 23.6 / 42.2 & 15.1 / 15.8 & 9.6 / 9.9 & 0.9 / 2.8 & 17.7 / 26.4 & 19.3 / 24.1 \\
    LLaVA-v1.6-34B & 25.5 / 27.7 & 65.5 / 68.4 & 28.3 / 47.2 & 22.8 / 48.7 & 16.8 / 18.1 & 10.8 / 13.4 & 3.7 / 10.3 & 16.3 / 30.5 & 20.2 / 28.3 \\
    CogVLM-Chat & 29.8 / 46.8  & 75.1 / 72.3 & 40.6 / 50.0 & 23.6 / 45.2 & 28.1 / 29.1 & 21.6 / 27.3 & 4.7 / 7.5 & 27.1 / 33.3 & 29.4 / 35.2 \\
    CogVLM2-Llama3-Chat & 34.0 / 31.9 & 73.4 / 75.7 & 46.2 / 49.1 & 24.7 / 41.1 & 28.4 / 32.2 & 26.8 / 32.5 & 5.6 / 8.4 & 30.2 / 40.0 & 31.4 / 37.9 \\
    InternVL2-26B & 40.4 / 44.7 & 77.4 / 79.7 & 48.1 / 54.7 & 27.8 / 54.0 & 27.4 / 31.5 & 24.5 / 27.8  & 3.7 / 15.0 & 36.7 / 44.1 & 32.7 / 40.2 \\ 
    Qwen2-VL-7B &  38.3 / 34.0 &	72.3 / 75.1 & 	48.1 / 48.1 &	19.4 / 53.2 &	25.1 / 23.8 &	20.0 / 20.2	& 4.7 / 9.3 &	29.5 / 33.3 &	28.4 / 33.1             \\ 
    LLaVA-OV-7B-Chat  & \colorbox{gray!20}{\textbf{42.6}} / 46.8 &	75.1 / 76.3 &	\colorbox{gray!20}{\textbf{50.9}} / 48.1 &	18.3 / 46.4  &	25.5 / 30.4 &	25.6 / 26.4 & 	9.3 / 17.8	&  33.8 / 42.4 & 	30.9 / 37.9\\ 
    GPT-4V & 40.4 / 40.4 & 74.0 / 70.1 & 44.3 / 49.1 & \colorbox{gray!20}{\textbf{34.2}}/57.4 & 28.0 / 33.2 & 25.4 / 29.2 & 10.3 / 18.7 & 42.7 / 51.6  & 34.8 / 41.8 \\  
    GPT-4o & 38.3 /\colorbox{gray!40}{\textbf{51.1}} & \textbf{\colorbox{gray!20}{79.7}/\colorbox{gray!40}{82.5}} & \textbf{\colorbox{gray!20}{50.9}/\colorbox{gray!40}{62.3}} & 29.3/\colorbox{gray!40}{\textbf{73.4}} & \textbf{\colorbox{gray!20}{42.2}/\colorbox{gray!40}{48.8}} & \textbf{\colorbox{gray!20}{42.1}/\colorbox{gray!40}{50.6}} & \textbf{\colorbox{gray!20}{11.2}/\colorbox{gray!40}{34.6}} & \textbf{\colorbox{gray!20}{43.2}/\colorbox{gray!40}{58.5}} & \textbf{\colorbox{gray!20}{42.7}/\colorbox{gray!40}{56.5}} \\  
    \hline
    Oracle  & 91.5 & 97.8 & 94.3 & 81.4 & 98.1 & 92.2 & 89.7 & 92.3 & 93.2 \\
    \hline
    \end{tabular}
    \caption{\label{tab:cogid}
    Cognition Scores of LVLMs on the Description task evaluated by GPT-4.
    The results of the evaluation under the Spontaneous Description mode and Directed Reasoning mode are presented before and after the ``/'' in each table cell. 
    For reference, the Cognition Scores of Oracle are calculated based on the annotated [Description] in CogBench dataset as an estimated upper bound. 
    Numbers are presented in \% with a full score of 100\%.
    }
\end{table*}

\subsubsection{Cognition}

\textbf{General analysis.}
Table \ref{tab:cogid} shows the Cognition Scores of LVLMs on the Description task.
GPT-4o achieves the best performance and there is a large performance gap between GPT-4o and other open-source models. 
Among open-source models, InternVL2-26B, LLaVA-OV-7B-Chat, CogVLM models and Qwen2-VL achieve relatively better performance, with some approaching that of GPT-4V. 
In terms of different reasoning capabilities, all LVLMs show better performance on [Location Reasoning] than others, probably because it is a kind of relatively lower-level reasoning.
Differently, for [Event Reasoning], [Event Relationship Reasoning], and [Next Moment Event Reasoning], most open-source LVLMs show lower performance. 
The Cognition Scores of some open-source LVLMs are only around or even lower than 10\% across these dimensions, indicating they almost do not understand the story in the images.
In contrast, GPT-4o shows significantly better performance in these three kinds of reasoning capabilities.
Besides, though GPT-4o achieves the best performance, there is also a large gap between its Cognition Scores and the Oracle scores.
This indicates that LVLMs still have a lot of room for development in terms of cognitive abilities.

\noindent \textbf{Analysis based on different evaluation modes.}
When the Directed Reasoning mode is applied, significant performance improvements in the LVLMs can be observed compared to the Spontaneous Description mode.
This suggests that current LVLMs, while potentially grasping some high-level semantic information in images, cannot spontaneously generate comprehensive descriptions through reasoning from various perspectives without appropriate prompts.
Another intriguing finding is that for some open-source LVLMs, shifting from the Spontaneous Description mode to the Directed Reasoning mode results in noticeable performance improvements across various dimensions, such as [Character Relationship Reasoning]; however, for [Event Reasoning] and [Event Relationship Reasoning], the enhancements are rather limited.
This further highlights their struggle to fully comprehend events in the images, with event-based reasoning being a key area where open-source LVLMs differ significantly from GPT-4o.

\noindent \textbf{Analysis of recognition and cognition abilities.}
The performance gap between GPT-4o and open-source LVLMs in terms of cognition is obviously larger than that of recognition.
This suggests that cognitive reasoning is a more advanced ability than recognition, and that these LVLMs still have significant room for development.

\subsubsection{Case Study} %
\label{sec:case_study}

\begin{figure*}[h]
    \centering
    \includegraphics[width=\textwidth]{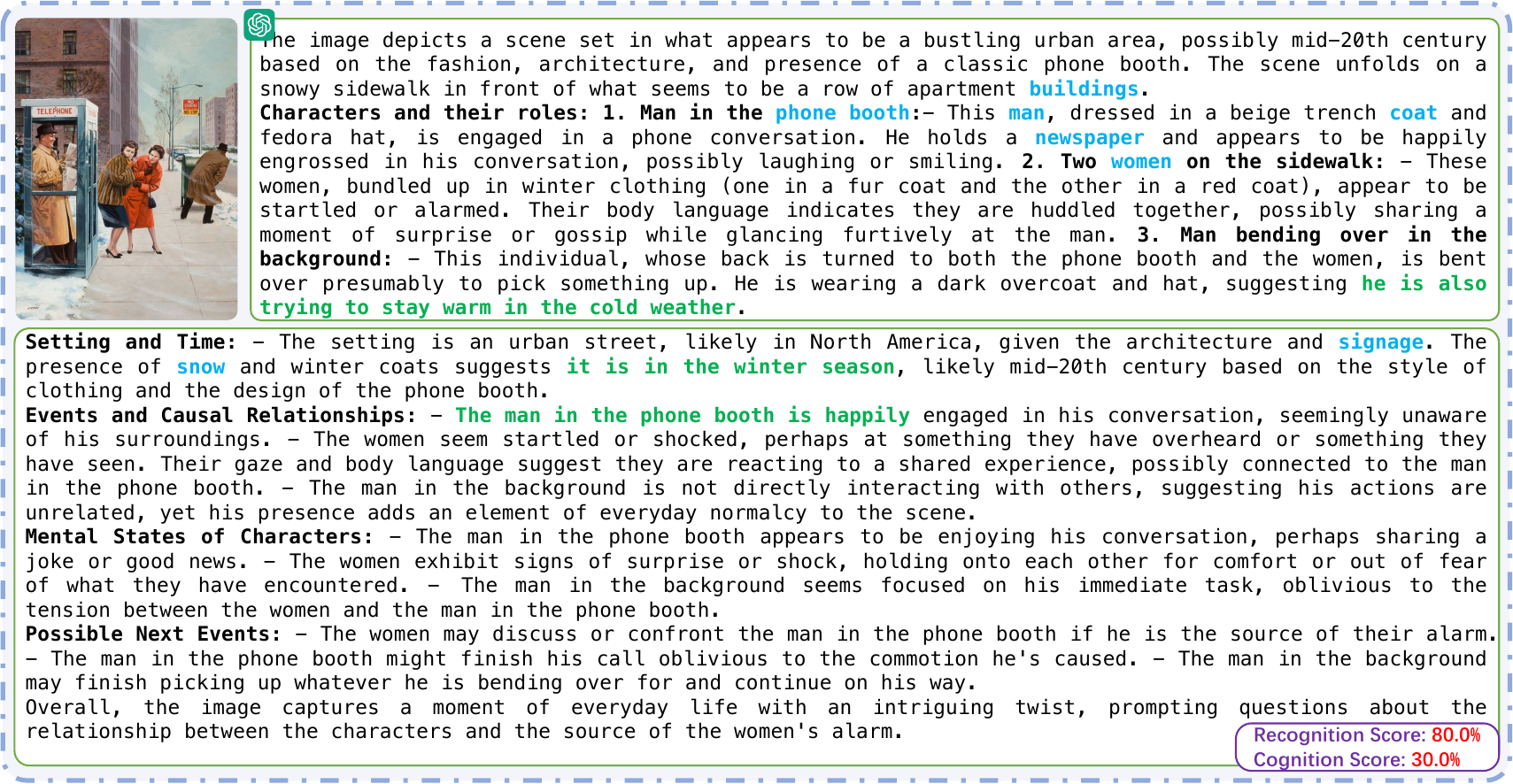}
    \caption{Case study of the Description task. 
    The description is generated by GPT-4o in the Directed Reasoning mode. 
    Recognized entities are marked in \textcolor{c3}{blue}, and CoRs are marked in \textcolor{c2}{green}.
    }
    \label{fig:case}
\end{figure*}

Figure \ref{fig:case} shows a failure case of GPT-4o on the Description task under the Directed Reasoning mode.
In terms of recognition, GPT-4o shows a good performance by recognizing most annotated entities such as \texttt{men, women, buildings, phone booth, newspaper, coat, snow, sign} and only fails to recognize \texttt{bus stop} and \texttt{wind}.  
However, GPT-4o fails to understand the story in the image and gets a 30.0\% in terms of cognition.
One of the most important reasons is that it does not recognize that the man is in the phone booth to escape the cold, rather than to make a phone call. 
This case demonstrates that CogBench reveals current LVLMs falling short in cognition, with a gap remaining between their cognitive abilities and human levels.

\subsubsection{Effectiveness of GPT-based Evaluation}

To validate the GPT-based cognition evaluation method, we manually scored CoRs of 20 images with a binary scale (0/1) and compared the accuracy of various evaluation methods on this subset. 
Table \ref{tab:eval_metrics} reveals that GPT-4 offers the highest accuracy, demonstrating that GPT-based evaluation aligns well with human assessment. 
Therefore, it is effective to assess LVLMs' performance on the Description task. 
Implementation details of evaluation methods beyond ChatGPT and GPT-4~\citep{lin2004rouge, bertscore, sellam2020bleurt, he2021deberta, yin-etal-2021-docnli} can be found in Appendix \ref{sec:eval_gpt_eval}. %

\begin{table}[h!]
    \centering
    \small

    \begin{tabular}{lc}
    \hline
    \thead{\textbf{Model}} & \thead{\textbf{Accuracy}}\\ 
    \hline
    ROUGE & 0.656 \\
    BERTScore & 0.635 \\
    BLEURT & 0.620 \\
    DeBERTa & 0.693 \\
    DocNLI  & 0.714 \\
    GPT-3.5 & 0.807 \\
    GPT-4 & 0.833 \\
    \hline
    \end{tabular}
    \caption{\label{tab:eval_metrics}
    CoR accuracy of cognition evaluation methods for the Description task.
     }
\end{table}

\subsection{Results of the VQA Task}

\begin{table*}[h!]
    \centering
    \scriptsize
    \setlength{\tabcolsep}{1.5pt} 

    \begin{tabular}{lccccccccc}
    \hline
    \thead{\textbf{Model}} & \thead{\textbf{Time} } & \thead{\textbf{Location} } & \thead{\textbf{Character} } &  \thead{\textbf{Character}\\ \textbf{Relationship} } & \thead{\textbf{Event} } & \thead{\textbf{Event} \\ \textbf{Relationship} } & \thead{\textbf{Next}\\ \textbf{Moment} \\ \textbf{Event} } & \thead{\textbf{Mental} \\ \textbf{State} } & \thead{\textbf{Overall} } \\ 
    \hline
    InstructBLIP-7B & 60.5 & 71.4 & 48.8 & 54.9 & 40.3 & 36.8 & 46.7 & 47.6 & 47.4 \\
    Qwen-VL-Chat & 65.1 & 82.3 & 60.5 & 54.3 & 50.9  &  45.0 & 47.4 & 51.1 & 54.0 \\
    LLaVA-V1.5-7B & 58.1 & 81.4 & 54.3 & 54.6 & 45.9 & 45.0  & 54.1 & 52.6  & 52.8 \\
    LLaVA-V1.5-13B & 69.8 & 82.3 & 65.4 & 59.9 & 50.2 & 47.3 & 57.8 & 57.1 & 57.3 \\
    mPLUG-Owl-2 & 51.2  & 81.8  & 58.6 & 54.6 &  46.0 &  47.5 & 47.4 & 51.8 & 52.7 \\
    ShareGPT4V-7B & 58.1 & 80.5 & 63.6 & 53.6 & 48.8  & 40.3  & 51.1 & 54.1 & 53.4 \\
    ShareGPT4V-13B & 67.4  & 80.0 & 65.4 & 56.5 & 49.4  & 49.8  & 60.0 & 54.6 & 56.3 \\
    LLaVA-v1.6-vicuna-7B  & 60.5 & 80.9 & 57.4 & 55.8 & 51.4 & 43.8 & 55.6 & 58.6 & 55.8 \\
    LLaVA-v1.6-vicuna-13B & 66.3 & 85.9 & 64.8 & 59.3 & 58.2 & 53.7 & 65.9 & 60.0 & 61.5 \\
    LLaVA-v1.6-34B & 80.2 & 92.7 & 83.3 & 74.8 & 68.4 & 66.9 & 68.8 & 74.2 & 73.7 \\
    CogVLM-Chat & 73.3 & 87.7 & 75.9 & 66.9 & 58.5 & 53.2 & 63.0 & 63.7 & 64.2 \\
    CogVLM2-Llama3-Chat & 73.3 & 92.3 & \colorbox{gray!30}{\textbf{86.4}} & 76.0 & 71.9 & 62.9 & 67.4 & 71.9 & 73.5 \\
    InternVL2-26B & \colorbox{gray!30}{\textbf{81.4}} & 91.8 & 82.7 & \colorbox{gray!30}{\textbf{78.2}} & 71.9 & 67.2 & 66.7 & 72.0 & 74.4 \\ 
    Qwen2-VL-7B & 80.2	& 90.0 &	82.1	& 75.1	& 66.4	& 63.2	&  72.6 & 69.8 & 71.6 \\
    LLaVA-OV-7B-Chat  & \colorbox{gray!30}{\textbf{81.4}}	& \colorbox{gray!30}{\textbf{93.2}}	 & 85.2	& 75.1	 &  71.6	& 69.4 & 	71.9 & 74.5 & 75.4  \\
    GPT-4V & 70.9 & 81.8 & 72.8 & 63.7 & 63.4  & 66.9  & 68.9 & 69.2 & 68.0 \\ 
    GPT-4o & \colorbox{gray!30}{\textbf{81.4}} & 90.9 & 81.5 & 68.5 & \colorbox{gray!30}{\textbf{75.5}}  &  \colorbox{gray!30}{\textbf{74.1}} & \colorbox{gray!30}{\textbf{83.7}} & \colorbox{gray!30}{\textbf{77.1}} & \colorbox{gray!30}{\textbf{77.1}} \\ 
    \hline
    Human & 98.8 & 95.9 & 98.8 & 94.3 & 95.6  & 96.0  & 96.3 & 93.3 & 95.3 \\
    \hline
    \end{tabular}
    \caption{\label{tab:cogvqa}
    Model performance on the VQA task. Each QA contains four options, with a chance rate of 25\%. 
    Numbers are presented in \% with a full score of 100\%.
    }
\end{table*}

Table \ref{tab:cogvqa} shows the performance of LVLMs on the VQA task. 
GPT-4o achieves the best performance.
Among open-source LVLMs, LLaVA-OV-7B-Chat, InternVL2-26B, LLaVA-v1.6-34B, CogVLM2, and Qwen2-VL-7B demonstrate better results, approaching the level of GPT-4o.

Consistent with previous findings, reasoning about location is also the easiest for LVLMs and event-related reasoning dimensions are more difficult.
There is also a large gap between the performance of LVLMs and humans.
Note that the accuracy of Human in Table \ref{tab:cogvqa} is calculated based on the responses of five healthy people. 
They all have obtained a bachelor's degree and are between the ages of 20 and 30.
Furthermore, some LVLMs perform differently on the two tasks, e.g. LLaVA-v1.6-34B, which highlights the necessity of including both tasks in the design of CogBench.

\section{Related Work}

\paragraph{Evaluation Benchmark for LVLMs.}
To better understand the emerging capabilities of LVLMs, various evaluation benchmarks have been proposed.
LVLM-eHub \cite{xu2023lvlm} evaluates LVLMs' capabilities across six categories using various publicly available CV datasets.
MME \cite{fu2023mme}, MMBench \cite{liu2023mmbench} and SEED Bench \cite{li2023seed} use True/False questions or multiple-choice questions to evaluate different abilities.
MM-VET \cite{yu2023mm} evaluates LVLMs in terms of their integrated VL capabilities with open-ended questions. 
In contrast, CogBench focuses on high-level cognitive evaluation. 
Though some of them also consider cognition as one of the evaluation dimensions, they do not provide a comprehensive evaluation and most images they use evaluate only limited aspects of reasoning.

\paragraph{Image Captioning.}
Image Captioning is a classical VL task~\cite{zhou2022can}. 
As model capabilities advance, researchers strive to enhance their ability to describe images in detail.
\citet{krause2017hierarchical} propose Image Paragraph Captioning, tasking models with generating a descriptive paragraph for an image.
Recently, some researchers \cite{xie2022visual, zhu2023chatgpt, zhuge2023mindstorms,chen2023sharegpt4v} have been leveraging the ability of LLMs to generate more detailed image descriptions.
None of these tasks considers evaluating the high-level cognitive abilities of models through description.
HL dataset \cite{cafagna2023hl} requires models to generate high-level captions, but it only considers three aspects (scene, action, rationale).
The content of most images in existing datasets does not reach the level of a story. 
This reveals the need for higher-level datasets like CogBench.

\paragraph{Visual Reasoning.}
Visual Reasoning task is closely related to the cognitive abilities of models.
Visual Commonsense Reasoning (VCR) \cite{zellers2019vcr} tasks models with answering visual questions using commonsense reasoning and justifying their answers.
VisualCOMET \cite{park2020visualcomet} is a framework of visual commonsense reasoning tasks to predict past, future events, and present intents.
\citet{hessel2022abduction} utilize images from VCR and Visual Genome \cite{krishna2017visual} to evaluate the ability of models to perform abductive reasoning.
\citet{FZCVR22} propose a task to identify the time and location of a given image.
CURE \cite{chen2023measuring} is proposed to measure both the zero-shot reasoning performance and consistency of VLMs. 
Similarly, these tasks consider fewer kinds of reasoning, and CogBench can be seen as the next step of these efforts.

\section{Conclusion}

In this paper, we incorporate the concept of the Cookie Theft picture description task into the evaluation of high-level cognitive abilities of LVLMs and design a novel evaluation benchmark called CogBench.
The images in CogBench are of high quality and require more complex cognitive reasoning for interpretation, distinguishing it from existing image datasets.
Experiments reveal a significant gap between the cognitive abilities of LVLMs and humans, highlighting CogBench as a challenging benchmark.

\section*{Limitations}

Given the scarcity of images that meet our data collection standards, the number of images in CogBench is relatively limited. 
However, the number of images in CogBench is comparable to existing evaluation benchmarks, such as MM-VET, and is sufficient to serve as a reliable evaluation benchmark.

\section*{Ethical Considerations}
\label{sec:ethical}

Most images in CogBench are manually collected from Pinterest in accordance with its terms of service.
The images are used under fair use for research purposes only and we will share our dataset with researchers who adhere to the ethical considerations established in this study.
During the annotation process, we ensure that our annotators receive a fair wage and promptly address any questions they have. 
They are free to take breaks or quit the annotation task at any time.

\section*{Acknowledgments}
This work has been supported by the China NSFC Project (No. U23B2018). 
Kenny Q. Zhu was partly supported by NSF Award No. 2349713.

\bibliography{acl}

\appendix

\clearpage
\section{Image Annotation Instruction}
\label{sec:annotation_instruction}
Figure \ref{fig:annotation} shows the image annotation instruction for annotators of CogBench.

\begin{figure*}
    \begin{tcolorbox}[
      colframe = blue!30!white, 
      colback = blue!2!white,
      colbacktitle = blue!10!white,
      colupper = black, collower = yellow!75!red,
      coltitle = black!90!white
      ]
      \small
      You are going to see some pictures. Each picture tells a story and requires different kinds of reasoning to fully understand it. You will be first asked to identify the entities and reasoning processes in the picture. Then, you will need to describe the story of the picture based on your identified entities and reasoning processes. \\
  
      First, you will be asked to identify the entities in the picture. The annotation format is [A, B, C], where A, B, C are entities. \\
  
      [Entities]: Please list the entities appearing in the picture, including people, animals, objects etc. You are encouraged to list as many entities as possible. Note that these entities need to be in your picture description afterwards. For entities that are difficult to recognize, please do not list them here or describe them. \\
  
      Then, you will be asked to identify different reasoning processes in the picture. The annotation format should follows the structure [A1 + A2 -> B], where A1 and A2 are premises and B is the conclusion. Note that if you write a conclusion, there must be at least one premise. Do not write a conclusion only, like [B]. Please write one conclusion at a time, and do not write a reasoning process like [A1 -> B -> C], which should be split into two. Each picture does not necessarily requires all kinds of reasoning. Please write None, if a picture does not involve a specific kind of reasoning or it is not important in the picture. \\
  
      [Special Time Reasoning] Please write your reasoning processes about the special time of the story in the picture, e.g. festivals, seasons etc. The special time is usually relevant to the story of the picture. For instance, if it is daytime in a picture, it is easily recognized, requires no reasoning and there is nothing special, you can write None. However, if there is a lamp on or a clock indicating a specific time, you can write down your reasoning about the time. \\
  
      [Location Reasoning] Please write your reasoning processes about the location of the story in the picture, e.g. near the school. \\
  
      [Character Reasoning]: Please write your reasoning processes about the characters of the subjects in the picture, e.g. a teacher, a doctor etc. \\
  
      [Character Relationship Reasoning]: Please write your reasoning processes about the relationships between the characters in the picture, e.g. mother-son relationship.  \\
  
      [Event Reasoning]: Please write your reasoning processes about the events in the current and previous moments in the picture based on the clues provided. Note that you only need to annotate those high-level events and can ignore the low-level ones. For instance, “the woman is looking at the man” is a low-level event and you can ignore its reasoning process. Differently, the reasoning process [A mother is busy cooking. + A boy is fetching cookies behind the mom. + A girl is shushing the boy. -> The boy is stealing cookies.] is about a high-level event ``stealing'' and you should write it down. \\
  
      [Event Relationship Reasoning]: Please write your reasoning processes about the relationships between different events in the picture. These events are usually linked through causal and temporal relations. Note that events in this part do not necessarily appears in the [Event Reasoning] part as some events here are low-level events. \\
  
      [Next Moment Event Reasoning]: Please write your reasoning processes about the events that will happen in the next moment. Note that you only need to write down events that have a very high probability of happening, instead of guessing what might happen next. \\
  
      [Mental State Reasoning]: Please write your reasoning processes about the mental states of the subjects in the picture, e.g. daydreaming, happy, etc. You need to reason as best you can about the mental states of all the subjects in the picture, unless they are not showing obvious emotions. \\
  
      Finally, you will be asked to describe the picture in as much detail as you can.\\
  
      [Description]: Please describe all you see in the picture in a paragraph based on the entities and reasoning processes you identified above, ensuring that all of them are included in your description. Each picture has a story behind it and you need to tell that story through your description. \\
    \end{tcolorbox}
    \caption{Image annotation instruction for annotators.}
    \label{fig:annotation}
\end{figure*}

\section{Prompt of CoR-based GPT-assisted Question Generation}
\label{sec:qa_prompt}

Figure \ref{fig:cgqg_prompt} shows an example prompt of the CoR-based GPT-assisted question generation method for GPT-4. 
This prompt is used to generate questions based on [Event Reasoning] CoRs. 
Prompts for other reasoning types are similar to this one.

\begin{figure*}[h]
    \small
    \begin{tcolorbox}[
      colframe = blue!30!white, 
      colback = blue!2!white,
      colbacktitle = blue!10!white,
      colupper = black, collower = yellow!75!red,
      coltitle = black!90!white,
      ]
    We have a description of an image and the description tells a detailed story unfolding in the image. In the process of describing an image, it is often necessary to engage in reasoning about events based on the clues within the image, leading to certain conclusions. For example, when we see the wind is blowing outside, and a man is reading a newspaper in the telephone booth, we can infer that he is actually hiding from the wind in the telephone booth. Therefore, in this task, in addition to the image description, the reasoning processes about event within the image description have also been extracted. For each reasoning process, we use A1+A2+...->B to represent it, where A1, A2, ... are clues we observed in the picture and B represents the conclusion about event we inferred. \\

    Thus, given an image description and the reasoning processes about event, our task is: \\
    1) Generate a question based on reasoning processes about event. \\
    2) Generate four options: A, B, C, and D. There is only one correct answer among the four options, which is consistent with the description and reasoning processes provided. The correct answer option should be randomly chosen from A, B, C, and D. For those incorrect options (distractors), you are encouraged to hallucinate some clues that are highly relevant to the question and the description but do not actually consistent with the description. That is, you can distort the facts in the description and reasoning processes using elements related to the question to generate some easily selectable distractors. It would be better if you can generate some distractors that are similar to but different from the correct option. Please avoid situations where the correct option is significantly longer or shorter than the distractors. \\
    --- \\
    For example, if the description is "There are some snow on the ground and it is windy, ... We can see it is cold. Inside a phone booth, a man is smiling while looking at a newspaper. He is sheltering from the cold wind in the phone booth..." and the question is "Why can we tell that the man is seeking shelter for warmth?", you can use "newsstand", which is related to "seeking shelter for warmth" in the question, to distort the fact in description "in a phone booth." Then you can get "the man is in the newsstand." Similarly, you can hallucinate a question related distractor "it is raining and a man is smiling and reading a newspaper in a phone booth," which is similar to the correct option "it is windy and a man is smiling and reading a newspaper in a phone booth," but different from it and inconsistent with the description.  \\
    --- \\
    3) Generate the the letter corresponding to the correct answer, that is A, B, C, or D.\\

    Here are some examples: \\
    ---------- \\

    [Description]:  \\
    There are some snow on the ground and it is windy, indicating it is winter. There are two men and two women standing on the roadside. There is a sign that says "NO STANDING BUS STOP", indicating it is near a bus stop. A man is standing on the road side, wrapping his coat tightly around himself, and peering out onto the road. They are probably waiting for a bus here. We can see it is cold. Inside a phone booth, a man is smiling while looking at a newspaper. He is sheltering from the cold wind in the phone booth. He looks happy, because it is warm there. Two women are also wrapping their coats tightly and looking at the man in the phone booth. They are probably friends and standing together. They are unhappy with the man. There are some buildings by the road. \\

    [Event Reasoning]:  \\ 
    It is windy and cold. + A man is standing in a phone booth reading newspaper. -> The man is sheltering from the cold wind in the phone booth. \\

    [Generated Multiple-Choice Questions]:  \\
    What is the man doing in the phone booth? \\
    A. Making a phone call. \\
    B. Reading a book. \\
    C. Avoiding someone he doesn't want to see. \\
    D. Sheltering from the cold wind. \\
    Correct Answer: [D] \\

    Why can we tell that the man is seeking shelter for warmth? \\
    A. It is windy and a man is smiling and reading a newspaper in a newsstand. \\
    B. It is raining and a man is smiling and reading a newspaper in a newsstand. \\
    C. It is windy and a man is smiling and reading a newspaper in a phone booth. \\
    D. It is raining and a man is smiling and reading a newspaper in a phone booth. \\
    Correct Answer: [C] \\
    ---------- \\ 
    Please: \\
    1). Generate at least one question for each reasoning process. \\
    2). Generate more diverse questions, try to generat questions from different perspectives or angles and don't limit yourself to the question templates provided in the examples. \\
    3). Avoid generating repetitive questions with similar meanings. \\
    \end{tcolorbox}
    \caption{An example prompt of CoR-based GPT-assisted question generation for GPT-4 to generate questions based on [Event Reasoning] CoRs.}
    \label{fig:cgqg_prompt}
  \end{figure*}

\section{Introduction to Selected LVLMs}
\label{sec:lvlms}

\begin{table}[h]
  \centering
  \scriptsize
  \setlength{\tabcolsep}{2.5pt}

  \begin{tabular}{lll}
  \hline
  \thead{\textbf{Model}} & \thead{\textbf{Visual Encoder} } & \thead{\textbf{Language Model}}\\ 
  \hline
  InstructBLIP-7B  & EVA-G & Vicuna-7B \\ 
  Qwen-VL-Chat & ViT-G/16 & Qwen-7B \\ 
  LLaVA-v1.5-7B  & CLIP ViT-L/14 & Vicuna-v1.5-7B \\
  LLaVA-v1.5-13B & CLIP ViT-L/14 & Vicuna-v1.5-13B \\
  mPLUG-Owl-2 & CLIP ViT-L/14 & LLaMA2-7B \\
  ShareGPT4V-7B & CLIP ViT-L/14 & Vicuna-v1.5-7B \\
  ShareGPT4V-13B & CLIP ViT-L/14 & Vicuna-v1.5-13B \\
  LLaVA-v1.6-vicuna-7B & CLIP ViT-L/14 & Vicuna-v1.5-7B \\
  LLaVA-v1.6-vicuna-13B & CLIP ViT-L/14 & Vicuna-v1.5-13B \\
  LLaVA-v1.6-34B & CLIP ViT-L/14 & Nous-Hermes-2-Yi-34B \\
  CogVLM-Chat & EVA2-CLIP-E & Vicuna-v1.5-7B \\
  CogVLM2-Llama3-Chat & EVA2-CLIP-E & LLaMA3-8B \\
  InternVL2-26B & InternViT-6B & InternLM2-Chat-20B \\
  Qwen2-VL-7B & QwenViT & Qwen2-7B \\
  LLaVA-OV-7B-Chat & SigLIP-400M & Qwen2-7B \\ 
  GPT-4V & \multicolumn{1}{c}{--}  & \multicolumn{1}{c}{--} \\
  GPT-4o & \multicolumn{1}{c}{--}  & \multicolumn{1}{c}{--} \\
  \hline
  \end{tabular}
  \caption{\label{tab:lvlms}
  LVLMs evaluated in this paper.
  }
\end{table}

\begin{itemize}
    \item 
    \textbf{InstructBLIP} \cite{dai2023instructblip} is an extension of BLIP-2 \cite{li2023blip}, designed to tackle the challenges of vision-language instruction tuning. 
    It consists of an image encoder, an LLM, and a Q-Former. 
    We use ``blip2\_vicuna\_instruct'' + ``vicuna7b'' for testing. 
    \item \textbf{Qwen-VL-Chat} \cite{bai2023qwenvl} is the instruction-tuned VL chatbot based on Qwen-VL, which consists of a visual encoder, an LLM, and a position-aware vision-language adapter. 
    Its training process consists of two pre-training stages followed by a final instruction fine-tuning stage. 
    We test ``Qwen-VL-Chat'' in the paper.
    \item \textbf{LLaVA v1.5} \citep{liu2023improved} is an upgraded version of LLaVA  \cite{liu2023llava}, who connects a vision encoder and LLM for visual and language understanding. 
    LLaVA is instruction-tuned on the language-image instruction-following data generated by language-only GPT-4. 
    By using CLIP-ViT-L-336px with an MLP projection and adding academic-task-oriented VQA data with simple response formatting prompts, LLaVA v1.5 achieves better performance. 
    ``llava-v1.5-7b'' and ``llava-v1.5-13b'' are tested.
    \item \textbf{mPLUG-Owl-2} \cite{ye2023mplug} leverages modality collaboration to enhance performance across both text and multi-modal tasks. 
    It adopts a modularized network design, with the language decoder serving as a universal interface to manage different modalities.
    We test ``mplug-owl2-llama2-7b'' in the paper. 
    \item \textbf{ShareGPT4V} \cite{chen2023sharegpt4v} follows the design of LLaVA v1.5. 
    It incorporates a large-scale resource featuring highly descriptive captions into both the pre-training and supervised fine-tuning phases.
    We test ``ShareGPT4V-7B'' and ``ShareGPT4V-13B''.
    \item \textbf{LLaVA v1.6} \cite{liu2024llavanext} maintains the minimalist design and data efficiency of LLaVA-v1.5.
    It enhances LLaVA-v1.5 with dynamic high resolution, data mixture, and scaling of the LLM backbone.
    We test ``llava-v1.6-vicuna-7b-hf'', ``llava-v1.6-vicuna-13b-hf'' and ``llava-v1.6-34b-hf'' in the paper. 
    \item \textbf{CogVLM} \cite{wang2023cogvlm} comprises a ViT encoder, an MLP adapter, a pretrained LLM, and a visual expert module. 
    Unlike the common shallow alignment method that maps image features to the input space of language model, CogVLM uses a trainable visual expert module in the attention and FFN layers to bridge the gap between the frozen pretrained language model and image encoder.
    ``cogvlm-chat-hf'' is tested in the paper.
    \item \textbf{CogVLM2}~\cite{hong2024cogvlm2} inherits the visual expert architecture from CogVLM.
    Differently, CogVLM2 uses a 2×2 downsampling module to increase input resolution while maintaining efficiency, with LLaMA3-8B as its backbone.
    The pre-training and post-training data are also improved in terms of diversity and quality. 
    ``cogvlm2-llama3-chat-19B'' is tested.
    \item \textbf{InternVL2} \cite{chen2023internvl} family adopts a progressive alignment training strategy, resulting in the first vision foundation model natively aligned with LLMs, and enabling efficient training of large models with limited resources. 
    It supports multimodal input using a single parameter set and provides multitask output, including images, bounding boxes, and masks. 
    By connecting the LVLM with multiple downstream task decoders, it can be generalized to many VL tasks. 
    We test ``InternVL2-26B'' in this paper.
    \item \textbf{Qwen2-VL}~\cite{Qwen2VL} is built upon the Qwen-VL architecture. 
    It introduces the Naive Dynamic Resolution mechanism, which allows the model to handle arbitrary image resolutions, dynamically adjusting the number of visual tokens. 
    It also incorporates Multimodal Rotary Position Embedding (M-RoPE), enabling the effective fusion of positional information across text, images, and videos. 
    ``Qwen2-VL-7B-Instruct'' is tested.
    \item \textbf{LLaVA-OneVision-Chat}~\cite{li2024llava} is an enhanced version of LLaVA-OneVision, with improvements achieved through preference alignment for better visual-chat capabilities. 
    LLaVA-OneVision is developed by integrating insights into data, models, and visual representations in the LLaVA v1.6 series. 
    ``llava-onevision-qwen2-7b-ov-chat'' is tested and we use LLaVA-OV-Chat to refer to LLaVA-OneVision-Chat in this paper.
    \item \textbf{GPT-4V} \citep{openai2023gpt4} is a powerful LVLM developed by OpenAI. The version of ``gpt-4-turbo'' is tested.
    \item \textbf{GPT-4o} \citep{openai2023gpt4} is currently one of the most powerful multimodal models. It is a single model trained end-to-end across text, vision, and audio. The version of ``gpt-4o'' is tested.
\end{itemize}

Table \ref{tab:lvlms} shows an overview of the designs of different LVLMs.

\section{Prompts for GPT-based Cognition Evaluation of the Description Task}
\label{sec:cogid_eval_prompt}

Figure \ref{fig:eval} and Figure \ref{fig:er_eval} show the prompts used for cognition evaluation of the Description task for GPT-4 (or ChatGPT).

\begin{figure}[h]
  \begin{tcolorbox}
      [
      colframe = blue!30!white, 
      colback = blue!2!white,
      colbacktitle = blue!10!white,
      colupper = black, collower = yellow!75!red,
      coltitle = black!90!white
      ]
      \noindent
      \small
      Given a <DESCRIPTION> and some <KEY POINT>s, please tell me if the <DESCRIPTION> explicitly presents the exact or similar semantics of each <KEY POINT>. The following points are required: \\

      1) Instead of reasoning about whether each <KEY POINT> is possibly correct based on the <DESCRIPTION>, you only need to determine whether the <DESCRIPTION> mentions the semantics in the <KEY POINT>. \\
      2) Do not overlook the semantics in the <DESCRIPTION> that are semantically equivalent to the <KEY POINT> but expressed in different ways. For instance, if the <DESCRIPTION> mentions "The woman is playing with her son...", we can tell it successfully includes semantics in the <KEY POINT> "The woman is the mother of the boy." \\
      3) If several possible scenarios are listed using `or' at a <KEY POINT>, you only need to determine whether one of these scenarios is mentioned in the <DESCRIPTION>. \\

      Assign a score of 0 or 1 to each <KEY POINT>, where 0 represents NO and 1 represents YES. \\

      {<DESCRIPTION>:} \\
      \textcolor{blue}{\{Description generated by a model.\}} \\

      {<KEY POINT>:} \\
      \textcolor{c2}{1. \{Annotated key point 1.\}} \\
      \textcolor{c2}{2. \{Annotated key point 2.\}} \\
      \textcolor{c2}{...} \\
      \textcolor{c2}{N. \{Annotated key point N.\}} \\

      Please write your answers in ``[ ]'' with 0 or 1 in the following format (number + square brackets): \\
      
      1. [1]  2. [0] \\
      
      Your answers to the \textcolor{c2}{\{N\}} <KEY POINT>(s) above: \\
      \textcolor{c2}{1. [ ]  2. [ ] ... N. [ ]}  \\
  \end{tcolorbox}
  \caption{Cognition evaluation prompt of reasoning types other than [Event Relationship Reasoning].} 
  \label{fig:eval}
\end{figure}

\begin{figure}[h!]
  \begin{tcolorbox}[
      colframe = blue!30!white, 
      colback = blue!2!white,
      colbacktitle = blue!10!white,
      colupper = black, 
      collower = yellow!75!red,
      coltitle = black!90!white]
      \small
      Given a <DESCRIPTION> and some <EVENT RELATIONSHIP>s, please tell me whether this <DESCRIPTION> clearly depicts the cause-and-effect relationships between events. \\

      The format of a <EVENT RELATIONSHIP> follows the structure "A1 + A2 + ... + An -> B", where A1, A2, ..., An and B are events. Events A1, A2, ..., An are the causes of event B, and event B is the result caused by events A1, A2, ..., An. The criteria for judgment lie in whether the <DESCRIPTION> mentions these events and clearly depicts the causal relationships between them. \\

      Assign a score of 0 or 1 to each <EVENT RELATIONSHIP>, where 0 represents NO and 1 represents YES. \\

      {<DESCRIPTION>:} \\
      \textcolor{blue}{\{Description generated by a model.\}} \\

      {<EVENT RELATIONSHIP>:} \\
      \textcolor{c2}{1. \{Annotated event relationship 1.\}} \\
      \textcolor{c2}{2. \{Annotated event relationship 2.\}} \\
      \textcolor{c2}{...} \\
      \textcolor{c2}{N. \{Annotated event relationship N.\}} \\

      Please write your answers in ``[ ]'' with 0 or 1 in the following format (number + square brackets): \\
      
      1. [1]  2. [0] \\
      
      Your answers to the \textcolor{c2}{\{N\}} <EVENT RELATIONSHIP>(s) above: \\
      \textcolor{c2}{1. [ ]  2. [ ] ... N. [ ]}  \\

  \end{tcolorbox}
  \caption{Cognition evaluation prompt of [Event Relationship Reasoning].} 
  \label{fig:er_eval}
\end{figure}

\begin{table*}[h]
  \centering
  \small
  \setlength{\tabcolsep}{2.5pt} 
  \begin{tabular}{lcccccc}
  \hline
  \textbf{Model} & \thead{\textbf{METEOR} } & \thead{\textbf{CIDEr} } & \thead{\textbf{BLEU-1} } &  \thead{\textbf{BLEU-2} } & \thead{\textbf{BLEU-3} } & \thead{\textbf{BLEU-4}} \\ 
  \hline
  InstructBLIP-7B & 0.130 / 0.183 & 0.043 / 0.003  &  0.255 / 0.218 & 0.127 / 0.104 &  0.063 / 0.049 & 0.033 / 0.024 \\
  Qwen-VL-Chat & 0.130 / 0.196 & 0.037 / 0.016 & 0.242 / 0.262 & 0.124 / 0.138 & 0.059 / 0.069 & 0.030 / 0.036 \\ 
  LLaVA-V1.5-7B & 0.146 / 0.182 & 0.054 / 0.020 &  0.309 / 0.275 & 0.158 / 0.138 & 0.076 / 0.065 & 0.037 / 0.032\\
  LLaVA-V1.5-13B & 0.146 / 0.176 & 0.051 / 0.018 &  0.312 / 0.274 & 0.160 / 0.137 & 0.076 / 0.066 & 0.037 / 0.034 \\
  mPLUG-Owl-2 & 0.132 / 0.184 & 0.035 / 0.012 &  0.260 / 0.235 & 0.126 / 0.116 & 0.057 / 0.053 & 0.027 / 0.025 \\
  ShareGPT4V-7B & 0.162 / 0.191 & 0.017 / 0.014 &  0.259 / 0.222 & 0.120 / 0.113 & 0.050 / 0.053 & 0.024 / 0.027 \\
  ShareGPT4V-13B & 0.165 / 0.184 & 0.024 / 0.015 &  0.278 / 0.254 & 0.129 / 0.126 & 0.055 / 0.059 &  0.026 / 0.030 \\
  LLaVA-v1.6-vicuna-7B  & 0.169 / 0.190 & 0.026 / 0.000 & 0.278 / 0.159 & 0.134 / 0.085 & 0.060 / 0.041 & 0.029 / 0.021 \\
  LLaVA-v1.6-vicuna-13B  & 0.172 / 0.195 & 0.023 / 0.000 & 0.283 / 0.166 & 0.137 / 0.090 & 0.061 /  0.043 & 0.030 / 0.021 \\
  LLaVA-v1.6-34B  & 0.172 / 0.199 & 0.026 / 0.000 & 0.279 / 0.173 & 0.134 / 0.092 & 0.060 / 0.044 & 0.030 / 0.021 \\
  CogVLM-Chat & 0.157 / 0.174 & 0.069 / 0.056 & 0.297 /  0.326 & 0.151 / 0.154 & 0.075 / 0.069 & 0.039 / 0.034 \\
  CogVLM2-Llama3-Chat & 0.176 / 0.191 & 0.059 / 0.032 & 0.356 / 0.257 & 0.183 / 0.127 & 0.091 / 0.059 & 0.049 / 0.029 \\
  InternVL2-26B & 0.210 / 0.202 &  0.008 / 0.000 & 0.250 / 0.166 & 0.136 / 0.091 & 0.070 / 0.044 & 0.037 / 0.023 \\ 
  Qwen2-VL-7B & 0.205 / 0.183 & 0.007 / 0.000 & 0.222 / 0.154 & 0.121 / 0.084 & 0.061 / 0.041 & 0.032 / 0.021 \\
  LLaVA-OV-7B-Chat & 0.198 / 0.192 & 0.009 / 0.000 & 0.245 / 0.138 & 0.118 / 0.072 & 0.051 / 0.033 & 0.024 / 0.016 \\ 
  GPT-4V & 0.189 / 0.191 & 0.013 / 0.000 &  0.250 / 0.153 & 0.113 / 0.071 & 0.047 / 0.028 & 0.022 / 0.011 \\ 
  GPT-4o & 0.182 / 0.189 & 0.070 / 0.000 & 0.332 / 0.145 & 0.161 / 0.074 & 0.077 / 0.033 & 0.039 / 0.016 \\ 
  \hline
  \end{tabular}
  \caption{\label{tab:cap_eval}
  Model performance on the Description task evaluated using traditional image captioning metrics.
  The results of the evaluation under the Spontaneous Description mode and Directed Reasoning mode are presented before and after the ``/'' in each table cell. 
  }
\end{table*}

\section{Evaluation of LVLMs on the Description Task Using Traditional Image Captioning Metrics}
\label{sec:cap_eval}

Table \ref{tab:cap_eval} shows the model performance on traditional image captioning evaluation metrics.
Following \citet{krause2017hierarchical}, we use METEOR \cite{banerjee-lavie-2005-meteor}, CIDEr \cite{vedantam2015cider}, BLEU-1, BLEU-2, BLEU-3, and BLEU-4 \cite{papineni2002bleu} to evaluate the model performance on the Description task.
Similar to the findings of \citet{zhu2023chatgpt}, it can be observed that traditional image captioning evaluation metrics are not quite suitable for evaluating the Description task. 
There are two possible reasons. 
The first possible reason is that image descriptions are longer and more flexible than traditional image captions.
The second possible reason is that our Description task requires evaluation metrics to place more emphasis on high-level semantics in the description.

\section{Implementation of Non-GPT-Based Cognition Evaluation Methods for the Description Task}
\label{sec:eval_gpt_eval}

The cognition evaluation of the Description task is to determine whether each CoR is mentioned in the description. 
Apart from ChatGPT or GPT-4, some other evaluation methods are also implemented to perform this classification task, as shown in Table \ref{tab:eval_metrics}.

For methods based on ROUGE \cite{lin2004rouge}, BERTScore \cite{bertscore}, and BLEURT \cite{sellam2020bleurt},
we first split the description into sentences and then use each CoR as the reference to calculate the (recall) score for each sentence compared to the CoR. 
Then, the highest score among all calculated scores is taken as the score of the CoR corresponding to the description.
Finally, the score is converted into 0 or 1 using a threshold. 

We also tried Natural Language Inference (NLI) models to perform the task. 
First, we use DeBERTa \cite{he2021deberta, he2023debertav} to perform a sentence-level NLI task similar to the method mentioned above. 
If there is at least one ``Entailment'' for all the sentences, the score of the CoR will be 1.
The model we adopt is \textit{mDeBERTa-v3-base-xnli-multilingual-nli-2mil7}.
The second NLI model we tried is DocNLI \cite{yin-etal-2021-docnli}, which can directly take the description and CoR as input and perform the classification task. 



\end{document}